\documentclass[11pt, a4paper, onecolumn, copyright]{ai}

\usepackage{multirow}
\usepackage[authoryear, sort&compress, round]{natbib}
\usepackage{graphicx}
\usepackage{xspace}
\usepackage{microtype}
\usepackage{subcaption}
\usepackage{booktabs}

\usepackage{listings} 
\usepackage[most]{tcolorbox}
\tcbuselibrary{listings}

\usepackage{textcomp}
\usepackage{hyperref}
\usepackage{amsmath}
\usepackage{amssymb}
\usepackage{mathtools}
\usepackage{amsthm}
\usepackage{enumitem}

\usepackage{pifont}
\definecolor{darkgreen}{RGB}{50,100,0}
\definecolor{darkred}{RGB}{200, 0, 0}

\newcommand{\eg}{e.g.\xspace}
\usepackage{xurl}
\usepackage{titletoc}

\definecolor{tableblue}{HTML}{F2F7FF}
\definecolor{headergray}{HTML}{444444}

\colorlet{DarkGreen}{green!50!black}
\colorlet{DarkRed}{red}

\DeclareRobustCommand{\github}{\raisebox{-1.5pt}{\includegraphics[height=1.05em]{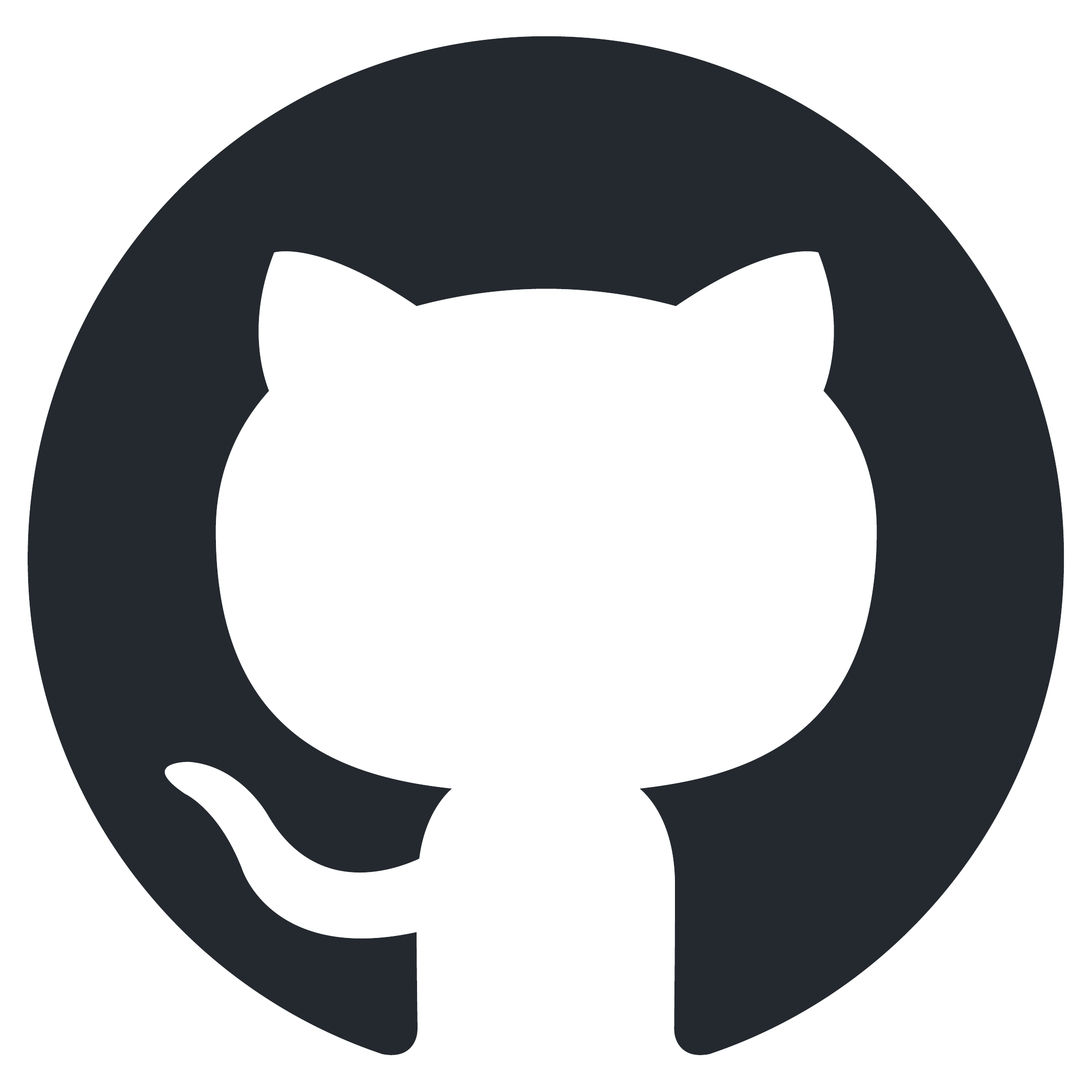}}\xspace}

\DeclareRobustCommand{\webpage}{\raisebox{-1.5pt}{\includegraphics[height=1.05em]{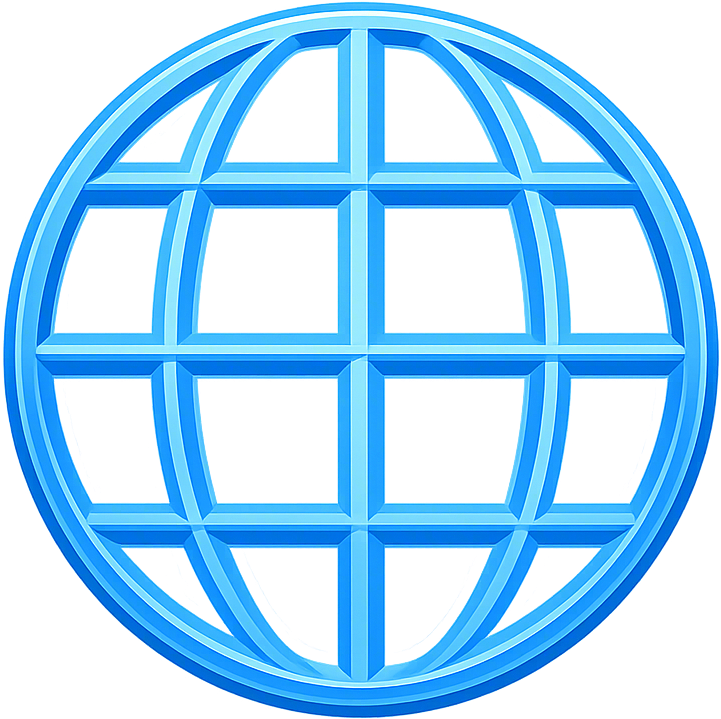}}\xspace}

\newcommand{\footerlinks}{%
  \github\ \href{https://github.com/lszhuhaichao/GravCal}{Repo}\hspace{1.5em}
  \webpage\ \href{https://lszhuhaichao.github.io/projects/image-based-gravity-estimation-vio-slam/}{WebPage}%
}

\newtcblisting{markdownbox}[1][]{
  listing engine=listings,
  listing options={
    basicstyle=\scriptsize\ttfamily, % 保持与你原来一致的 \scriptsize
    breaklines=true,              % 自动换行
    breakatwhitespace=true,
    numbers=none,                 % 不显示行号
    columns=fullflexible,         % 优化对齐
    keepspaces=true,              % 保留代码中的空格
    aboveskip=0pt,
    belowskip=0pt
  },
  colback=gray!5!white,           % 背景色：极浅灰
  colframe=gray!70!black,         % 边框色：深灰
  title={\textbf{Prompt (Markdown)}}, % 默认标题
  fonttitle=\bfseries,
  listing only,                   % 只显示代码，不运行
  left=5mm, right=5mm, top=3mm, bottom=3mm, % 内边距
  breakable,                      % 允许跨页
  enhanced,                       % 启用高级外观
  overlay={\begin{tcbclipinterior}\fill[gray!5!white] (frame.south west) rectangle ([xshift=0mm]frame.north west);\end{tcbclipinterior}}, % 纯净背景
  #1                              % 允许修改参数
}

\newtcbinputlisting{\markdownfile}[2][]{
  listing engine=listings,
  listing file={#2}, 
  listing options={
    basicstyle=\scriptsize\ttfamily,
    breaklines=true,
    breakatwhitespace=true,
    numbers=none,
    columns=fullflexible,
    keepspaces=true
  },
  colback=gray!5!white,
  colframe=gray!70!black,
  title={\textbf{Prompt (Markdown)}},
  fonttitle=\bfseries,
  listing only,
  breakable,
  enhanced,
  #1
}

\uselogo{}

\title{GravCal: Single-Image Calibration of IMU Gravity Priors with Per-Sample Confidence}

\renewcommand{\today}{}
\newcommand{\publicday}{Mar.~19, 2026}

\author[1 \hspace{-0.3em}$^\dag$]{Haichao Zhu}
\author[2]{Qian Zhang}

\correspondingauthor{lszhuhaichao@gmail.com}

\affil[1]{Independent Researcher}
\affil[2]{UC Riverside \footnote{$^\dag$Corresponding author.\hfill\textbf{Date:} \publicday.
}
}

\begin{abstract}

Gravity estimation is fundamental to visual-inertial perception, augmented reality, and robotics, yet gravity priors from IMUs are often unreliable under linear acceleration, vibration, and transient motion. Existing methods often estimate gravity directly from images or assume reasonably accurate inertial input, leaving the practical problem of correcting a noisy gravity prior from a single image largely unaddressed.

We present GravCal, a feedforward model for single-image gravity prior calibration. Given one RGB image and a noisy gravity prior, GravCal predicts a corrected gravity direction and a per-sample confidence score. The model combines two complementary predictions, including a residual correction of the input prior and a prior-independent image estimate, and uses a learned gate to fuse them adaptively.

Extensive experiments show strong gains over raw inertial priors: GravCal reduces mean angular error from $22.02^\circ$ (IMU prior) to $14.24^\circ$, with larger improvements when the prior is severely corrupted. We also introduce a novel dataset of over 148K frames with paired VIO-derived ground-truth gravity and Mahony-filter IMU priors across diverse scenes and arbitrary camera orientations. 
The learned gate also correlates with prior quality, making it a useful confidence signal for downstream systems.

\centerline{\footerlinks}
\end{abstract}

\makeatletter
\begin{document}

% --- hide footnote marker ONLY for title/affil footnote ---
\begingroup
\makeatletter
\renewcommand{\thefootnote}{}% no a/1/... marker
\renewcommand{\@makefnmark}{}% no mark in text
\maketitle
\makeatother
\endgroup

\begin{figure}[h]
    \centering
    \includegraphics[width=0.30\linewidth]{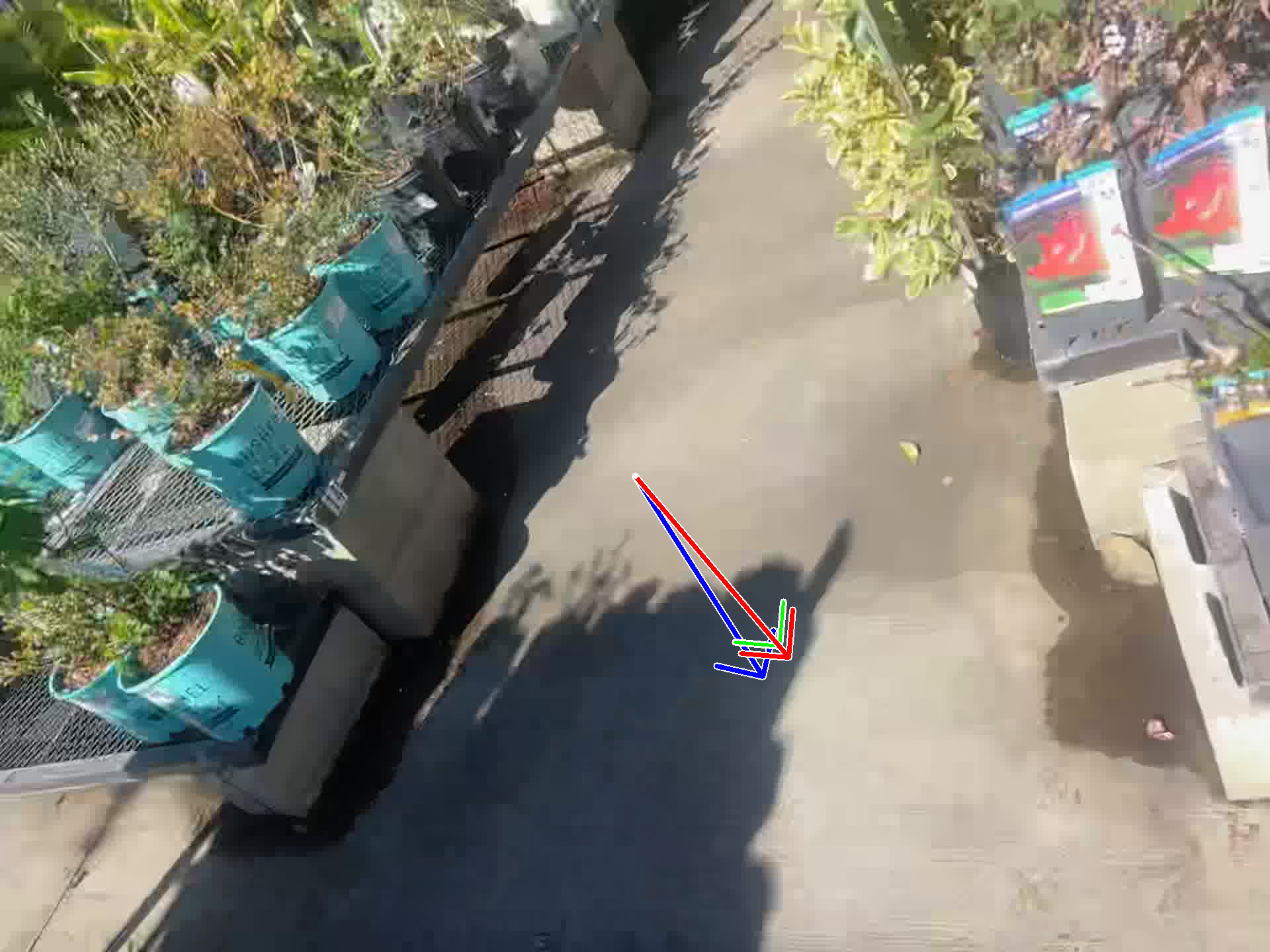}\hspace{0.3em}
    \includegraphics[width=0.30\linewidth]{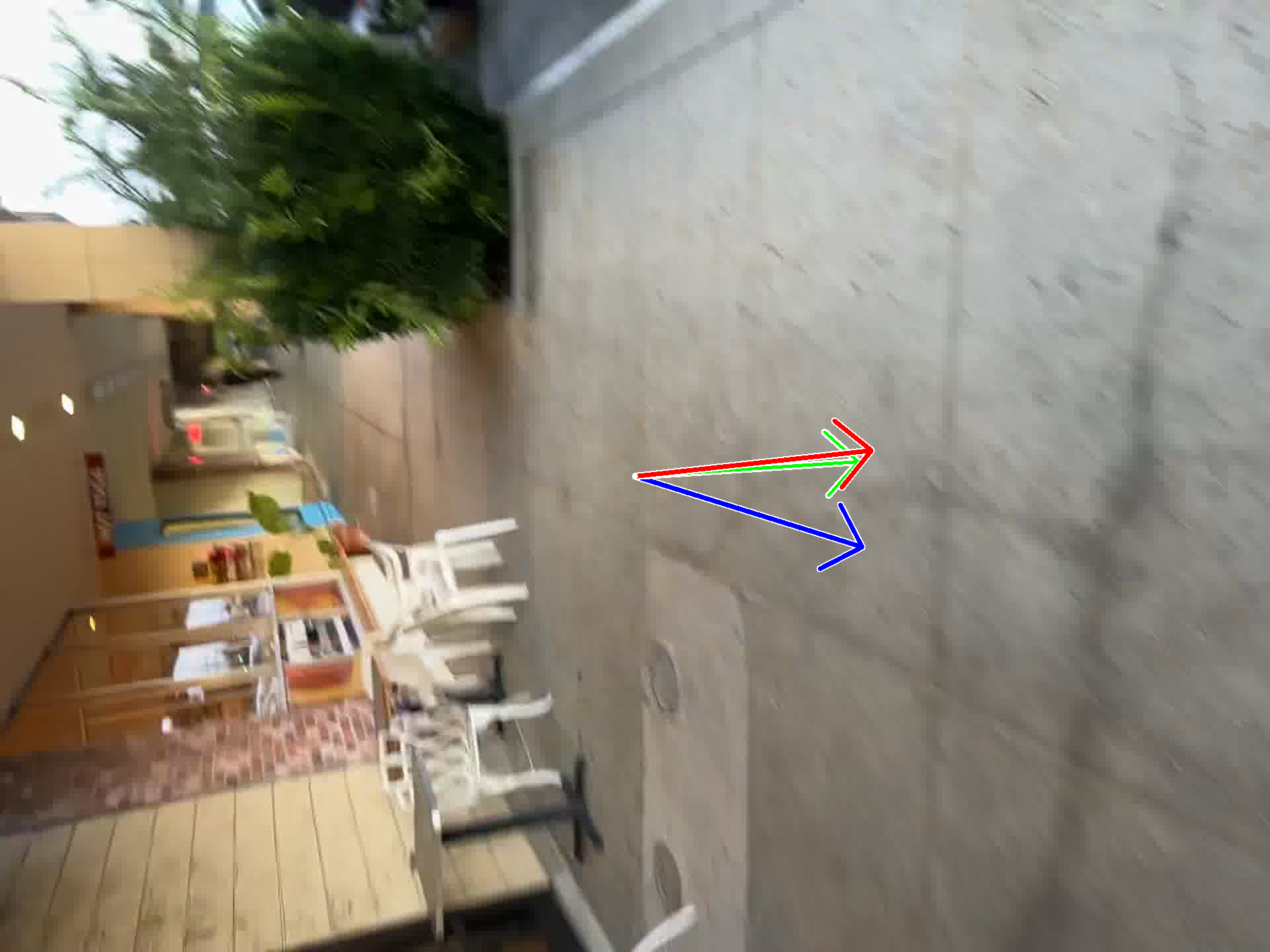}\hspace{0.3em}
    \includegraphics[width=0.30\linewidth]{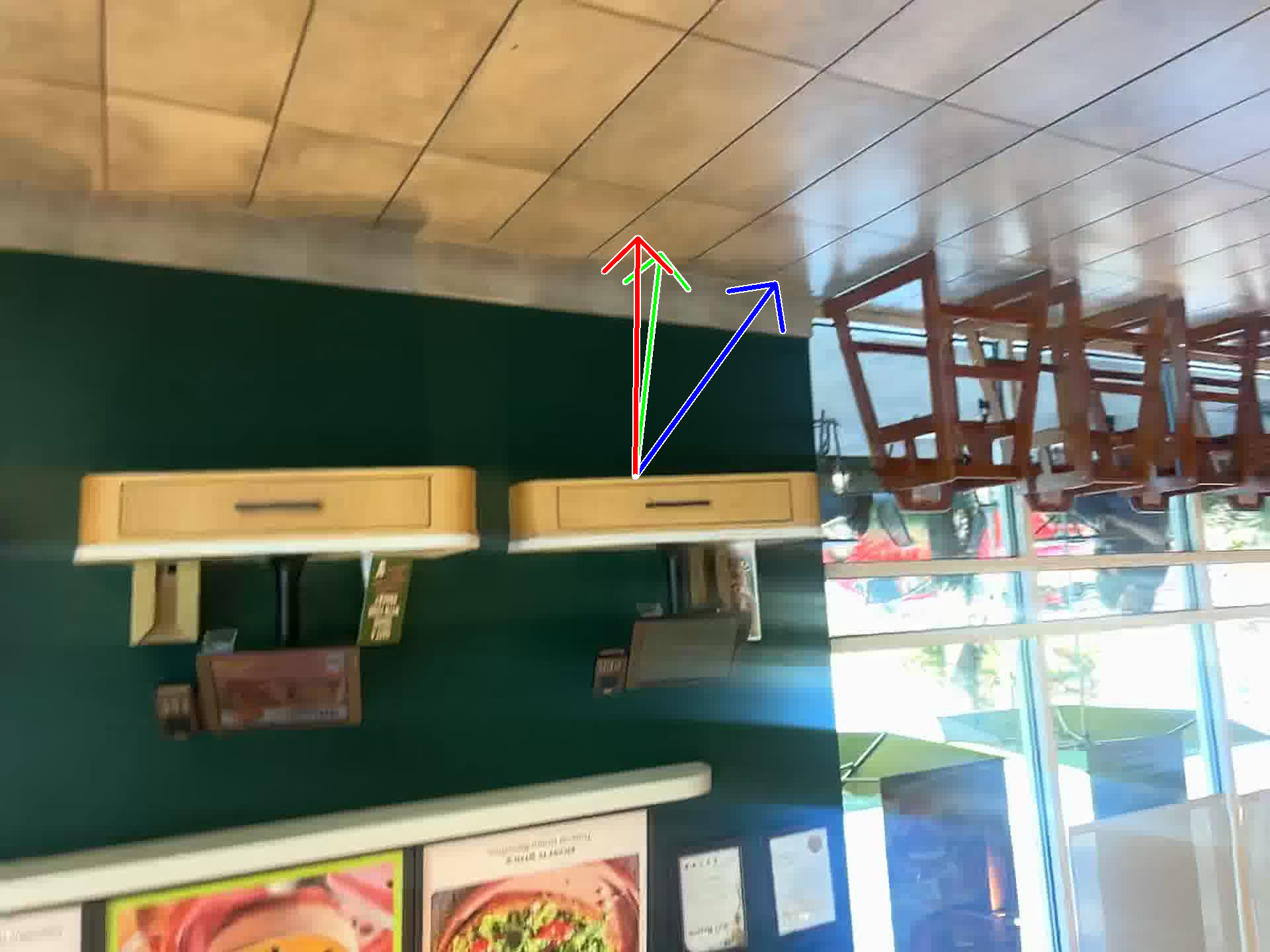}
    \par\smallskip
    \small
    \textcolor{blue}{$\boldsymbol{\rightarrow}$}~IMU prior\quad
    \textcolor{red}{$\boldsymbol{\rightarrow}$}~Ground truth\quad
    \textcolor{green}{$\boldsymbol{\rightarrow}$}~GravCal output
    \caption{\textbf{GravCal calibrates an IMU-derived gravity prior using a single image, at arbitrary camera orientations.}
        Examples span sideways ($90^\circ$) to near-inverted ($180^\circ$) poses.
        GravCal reduces the prior error in a single forward pass, without any orientation-specific tuning.
        }
    \label{fig:teaser}
\end{figure}

\section{Introduction}\label{sec:intro}

Knowing the gravity direction in the camera coordinate frame is a key prerequisite for many 3D vision tasks.
In visual localization, gravity constrains the camera pose to a single degree of freedom~\cite{mourikis2007multi,qin2018vins}, significantly reducing the search space.
In augmented reality, virtual objects must be aligned with the physical ground plane to appear realistic~\cite{fuvattanasilp2021slidar+}.
In robotics, gravity provides a global reference for navigation and manipulation~\cite{grip2013nonlinear}.
Despite its importance, estimating gravity from a single image remains a challenging problem, especially when the camera orientation is unconstrained.

Figure~\ref{fig:teaser} illustrates this setting: across sideways and near-inverted views, the IMU prior can deviate noticeably from the ground truth, while visual calibration pulls the estimate closer to the correct gravity direction.

Existing approaches to gravity estimation have complementary weaknesses under unconstrained real-world capture.
\emph{Inertial methods} use accelerometer and gyroscope signals via complementary filters~\cite{mahony2008nonlinear} or full visual-inertial systems~\cite{mourikis2007multi,qin2018vins}.
They are scene-agnostic and high-rate, but their gravity output degrades under dynamics: linear acceleration and vibration corrupt the accelerometer reference, while gyroscope drift accumulates over time, yielding heavy-tailed tilt errors.
\emph{Image-based methods} detect geometric cues~\cite{lee2009geometric,tardif2009non} or regress gravity from appearance~\cite{holdgeoffroy2018perceptual,veicht2024geocalib}.
They ignore inertial information available on modern devices and often rely on upright-biased visual statistics, which degrade under sideways or near-inverted poses.
Most importantly, both families output only a point estimate and provide no calibrated reliability signal, so downstream modules cannot decide when to trust or reject a prediction.

\begin{figure*}[t]
    \centering
    \scalebox{0.95}{
    \includegraphics[width=1\textwidth]{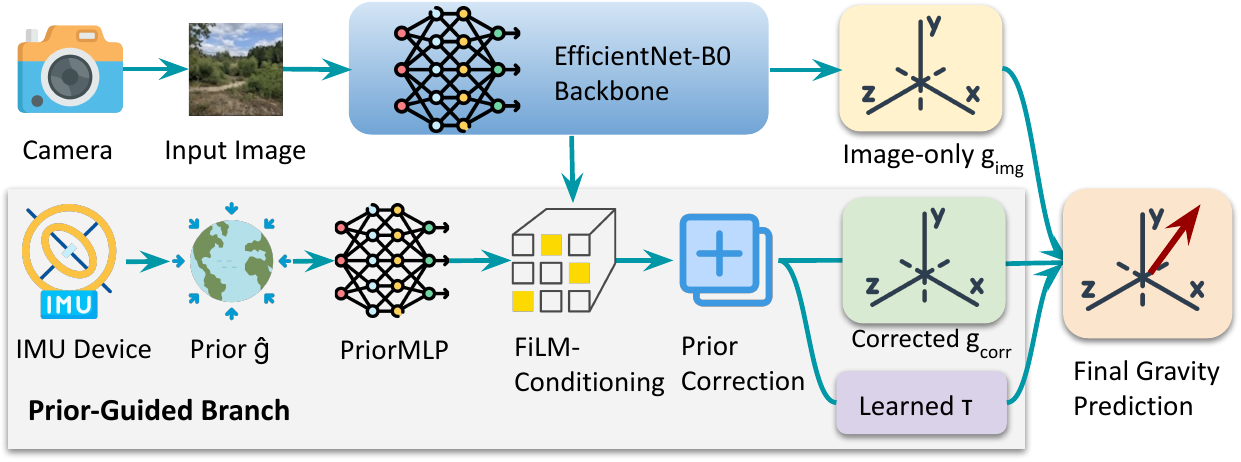}
    }
    \caption{\textbf{GravCal Overview.} 
    The input image is encoded by an EfficientNet-B0 backbone.
    In the prior-guided branch, the IMU prior $\hat{\mathbf{g}}$ is mapped by a PriorMLP and used to FiLM-condition the visual feature, which is then refined by a prior-correction module to produce the corrected estimate $\mathbf{g}_{\mathrm{corr}}$.
    In parallel, the backbone feature is also used to regress an image-only estimate $\mathbf{g}_{\mathrm{img}}$.
    The final gravity $\mathbf{g}_{\mathrm{pred}}$ is obtained by fusing the corrected prior and the image-only estimate via adaptive gating $\tau$.}
    \label{fig:architecture}
\end{figure*}

In this paper, we study \emph{single-image gravity prior calibration}: given a gravity prior derived from an IMU and a single RGB image, correct the prior \emph{and} produce a per-sample confidence score that signals when the result can be trusted.
IMU-based gravity estimates are readily available on modern devices but suffer from gyroscope drift and transient accelerometer corruption, producing heavy-tailed errors that propagate into downstream systems.
We present \textbf{GravCal}, a feedforward calibrator that addresses both goals in one forward pass.
As shown in Figure~\ref{fig:architecture}, GravCal conditions visual features on the input prior via FiLM~\cite{perez2018film} and produces two complementary hypotheses: a residual-corrected prior obtained by rotating $\hat{\mathbf{g}}$, and a prior-independent gravity estimate regressed directly from image features.
A learned gating scalar $\tau \in [0,1]$ adaptively fuses these branches---when $\tau \to 0$ the model trusts inertial correction; when $\tau \to 1$ it falls back on the image branch---and serves as an explicit per-sample reliability signal.

In addition to the model, we contribute a large-scale dataset for gravity prior calibration.
We collect over 148K images using iOS devices, spanning offices, natural outdoor environments, and indoor-outdoor transitions, with deliberate coverage of large camera rotations and motion-blur conditions.
Unlike existing datasets that predominantly contain near-upright images, ours explicitly covers a broad range of orientations---including sideways and near-inverted poses---together with paired visual-inertial odometry (VIO)-derived ground-truth gravity and IMU-derived Mahony-filter priors, enabling training and evaluation under unconstrained real-world conditions.

We summarize our contributions as follows:
\begin{itemize}
    \item We present \textbf{GravCal}, a single-pass model that jointly performs IMU prior calibration and per-sample reliability estimation under arbitrary camera orientation.
    \item We introduce a large-scale dataset of over 148K images spanning offices, outdoor and natural scenes, and indoor-outdoor transitions, with large camera rotations and motion blur, dually annotated with VIO-derived ground-truth gravity and IMU-derived Mahony-filter priors.
    \item Extensive evaluation demonstrates that GravCal reduces mean tilt error from $22.02^\circ$ (IMU prior) to $14.24^\circ$, and shows that the predicted gate $\tau$ correlates with IMU prior quality, providing a meaningful per-sample reliability signal for downstream use.
\end{itemize}

\section{Related Work}\label{sec:related}

\paragraph{Geometric and inertial gravity estimation.}
Classical geometric methods infer gravity from image structure, such as vanishing points and horizon cues~\cite{tardif2009non,lee2009geometric,zhai2016vp,hedau2009recovering}. They can perform well in structured scenes, but degrade when line cues are weak, cluttered, or occluded. IMU-based methods, including complementary filtering~\cite{mahony2008nonlinear} and visual-inertial estimators~\cite{mourikis2007multi,qin2018vins}, are scene-agnostic and high-rate, but become unreliable under strong dynamics: linear acceleration and vibration corrupt the accelerometer reference, while gyroscope drift accumulates over time.
Full VIO~\cite{mourikis2007multi,qin2018vins} jointly refines gravity alongside camera poses over time, but requires a well-initialized gravity estimate to converge, fails at cold start and after track loss, and is unavailable in single-frame settings such as visual relocalization.
GravCal is therefore complementary to VIO: it provides a fast, single-pass initialization and per-sample confidence estimate that can bootstrap or rescue any downstream pipeline.

\paragraph{Learning-based calibration and gravity prediction.}
Recent learning-based approaches estimate camera calibration or orientation from a single image~\cite{bogdan2018deepcalib,holdgeoffroy2018perceptual,veicht2024geocalib}. GeoCalib~\cite{veicht2024geocalib}, for example, combines deep features with iterative optimization to recover camera parameters including the up-vector. These methods are image-only and often rely on upright-biased visual statistics, which can fail under sideways or near-inverted viewpoints. In contrast, our setting explicitly uses the inertial prior available on commodity devices and focuses on calibrating that prior rather than discarding it.

\paragraph{Conditioning with physical priors.}
FiLM~\cite{perez2018film} is a general feature-conditioning mechanism widely used in vision models~\cite{huang2017adain,karras2019stylegan,goyal2017modulating}. We adopt FiLM as a physics-aware conditioning operator: instead of semantic codes, we condition on IMU-derived gravity. %This lets the network adapt its representation to the expected direction of gravity before regression, improving correction when the prior is informative.

\paragraph{Reliability-aware prediction.}
Prior-guided estimation is central in visual-inertial navigation~\cite{mourikis2007multi,qin2018vins}, but most gravity estimators output only point predictions. This is a practical gap: downstream systems need to know \emph{when} a prediction is trustworthy. GravCal addresses this by jointly predicting calibrated gravity and a gate $\tau$ that measures reliance on inertial versus visual evidence.
%yielding an explicit per-sample reliability signal for risk-aware downstream use.

\section{Dataset}\label{sec:dataset}

A key contribution of this work is a large-scale dataset for gravity estimation under unconstrained camera orientations.
Existing datasets used for camera calibration and gravity prediction predominantly contain images captured in near-upright poses, providing limited coverage of large rotations.
To address this gap, we collect a diverse dataset of over 148K images with ground-truth gravity annotations spanning the full range of camera orientations.
Each sample is annotated with both a precise gravity label derived from visual-inertial odometry (VIO) and a coarser IMU-based gravity prior, enabling us to train and evaluate models that leverage inertial measurements at inference time.

\subsection{Data Collection}\label{sec:data_collection}

We use the Stray Scanner application~\cite{strayscanner} on iOS devices to record RGB video sequences together with 6-DoF camera poses estimated by ARKit's visual-inertial odometry (VIO) system.
Each recording session produces an RGB video stream and a synchronized odometry file containing per-frame camera poses represented as quaternions and translation vectors.
During recording, we deliberately vary the device orientation---including upright, tilted, sideways, and upside-down poses---to ensure broad coverage of the orientation space.
Recordings are performed across diverse indoor and outdoor environments, covering offices, homes, streets, parks, and public spaces.

\subsection{Gravity Label Extraction}\label{sec:gravity_extraction}

ARKit defines a Y-up world coordinate system in which the gravity direction is $\mathbf{g}_w = (0, -1, 0)^{\top}$.
For each frame, the VIO system provides the camera-to-world rotation $R_{wc}$.
We compute the gravity direction in the camera coordinate frame as:
\begin{equation}
    \mathbf{g}_c = R_{wc}^{\top}\, \mathbf{g}_w = R_{cw}\, \mathbf{g}_w\,,
    \label{eq:gravity_cam}
\end{equation}
where $R_{cw} = R_{wc}^{\top}$ is the world-to-camera rotation.
The resulting vector $\mathbf{g}_c$ is a unit vector on $S^2$ indicating the direction of gravity as seen from the camera.

To ensure consistency with standard conventions used in visual-inertial navigation, we convert from the ARKit camera frame (X-right, Y-down, Z-forward) to the EuRoC convention (X-right, Y-forward, Z-up) via the coordinate permutation $(g_x, g_y, g_z) \mapsto (g_x, g_z, -g_y)$.

\subsection{IMU Gravity Prior}\label{sec:imu_prior}

In addition to VIO-derived ground-truth labels, we extract a per-frame gravity prior from the raw IMU stream recorded by Stray Scanner.
This prior mimics the signal available to a lightweight inference system that has access to inertial measurements but not a full VIO pipeline.

\paragraph{Mahony complementary filter.}
We process the synchronized accelerometer and gyroscope readings with a Mahony complementary filter~\cite{mahony2008nonlinear}.
The filter maintains a unit quaternion representing the orientation of the IMU body frame relative to a Z-up world frame, updated at the native IMU rate.
The accelerometer provides a low-frequency correction that counteracts gyroscope drift, while the gyroscope drives fast-rate integration.
At each frame, the estimated gravity direction in the IMU body frame is obtained as $\mathbf{g}_{\text{imu}} = -\hat{\mathbf{v}}_{\text{up}}$, where $\hat{\mathbf{v}}_{\text{up}}$ is the estimated ``up'' direction expressed in the body frame.

\paragraph{Frame alignment via Orthogonal Procrustes.}
The IMU body frame is mechanically fixed to the device but generally differs from the ARKit camera frame by an unknown constant rotation $R_{\text{imu}\to\text{cam}}$.
We recover this rotation by solving an Orthogonal Procrustes problem:
\begin{equation}
    R_{\text{imu}\to\text{cam}} = \arg\min_{R \in SO(3)} \sum_{i} \|\mathbf{g}^{(i)}_{\text{cam}} - R\,\mathbf{g}^{(i)}_{\text{imu}}\|^2,
    \label{eq:procrustes}
\end{equation}
where $\mathbf{g}^{(i)}_{\text{cam}}$ is the VIO ground-truth gravity for frame $i$ and $\mathbf{g}^{(i)}_{\text{imu}}$ is the Mahony-filter estimate.
The solution is obtained via SVD of the cross-covariance matrix $M = G_{\text{cam}}^{\top} G_{\text{imu}}$.
The aligned prior for all frames is then $\mathbf{g}^{\text{prior}}_{\text{cam}} = R_{\text{imu}\to\text{cam}}\,\mathbf{g}_{\text{imu}}$, expressed in the EuRoC camera convention.

\paragraph{Training setup.}
During training, the IMU-derived prior $\mathbf{g}^{\text{prior}}$ serves as the network's inertial input, and the VIO-derived gravity $\mathbf{g}_{\text{gt}}$ serves as the supervision target.
All training sequences include synchronized IMU data, so real Mahony-filter priors are available for every sample.

\subsection{Dataset Statistics}\label{sec:data_stats}

The final dataset comprises over 148K images extracted from 84 recording sessions.
Table~\ref{tab:dataset_stats} summarizes both split-level statistics and orientation distribution.

\begin{table}[t]
    \centering \small
    \caption{Dataset statistics and orientation distribution. Orientation is measured by tilt angle $\theta=\arccos(g_z)\in[0^\circ,180^\circ]$ in the camera frame, and grouped into three bins.}
    \label{tab:dataset_stats}
    \begin{minipage}[t]{0.47\linewidth}
        \vspace{0pt}
        \centering
        \textbf{Split Statistics}\\[0.2em]
        \setlength{\tabcolsep}{4.5pt}
        \begin{tabular}{lrr}
            \toprule
            Split & \#Sessions & \#Images \\
            \midrule
            Train & 53 & 95251 \\
            Val   & 12 & 26899 \\
            Test  & 19 & 26446 \\
            \midrule
            Total & 84 & 148K+ \\
            \bottomrule
        \end{tabular}
    \end{minipage}\hfill
    \begin{minipage}[t]{0.47\linewidth}
        \vspace{0pt}
        \centering
        \textbf{Orientation Distribution}\\[0.2em]
        \setlength{\tabcolsep}{4.5pt}
        \begin{tabular}{lrr}
            \toprule
            Tilt bin & \#Images & Ratio (\%) \\
            \midrule
            $0^\circ$--$60^\circ$     & 46738    & 31.47 \\
            $60^\circ$--$120^\circ$   & 85372    & 57.48 \\
            $120^\circ$--$180^\circ$  & 16425    & 11.06 \\
            \midrule
            Total       & 148K+ & 100.0 \\
            \bottomrule
        \end{tabular}
    \end{minipage}
\end{table}

\begin{figure}[t]
    \centering
    \includegraphics[width=\linewidth]{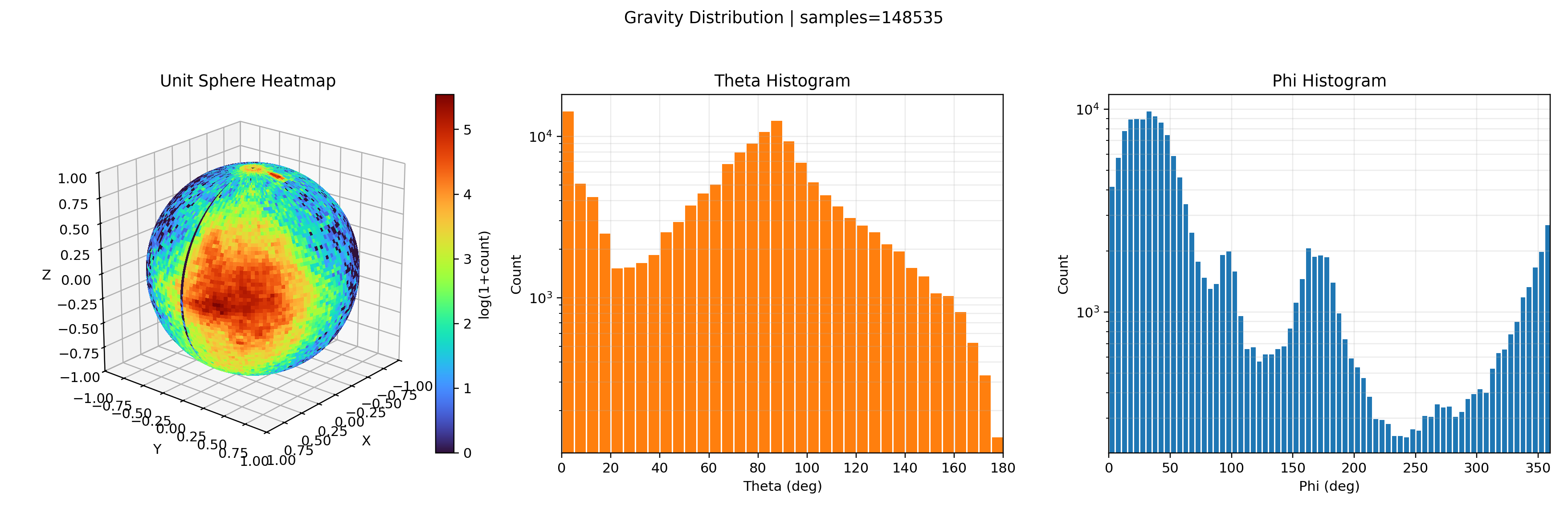}
    \caption{Distribution of gravity directions. The sphere heatmap shows directional density on $S^2$, and the marginal plots summarize angular coverage. The distribution spans a broad range of orientations rather than concentrating around upright poses.}
    \label{fig:gravity_distribution}
\end{figure}

Unlike datasets commonly used for gravity or camera calibration tasks, our dataset exhibits a broad distribution of gravity directions.
Figure~\ref{fig:gravity_distribution} visualizes the distribution of gravity vectors projected onto the unit sphere, confirming that our data covers a wide range of orientations.

\section{Method}\label{sec:method}

As shown in Figure~\ref{fig:architecture}, given a single undistorted RGB image $\mathbf{I} \in \mathbb{R}^{H \times W \times 3}$ and an IMU-derived gravity prior $\hat{\mathbf{g}} \in S^2$, our goal is to estimate the true gravity direction $\mathbf{g}^* \in S^2$ in the camera coordinate frame.
Inertial sensors provide a gravity estimate that is readily available on modern devices, but the Mahony complementary filter~\cite{mahony2008nonlinear} output suffers from gyroscope drift and residual calibration errors, resulting in a prior that is directionally informative yet not sufficiently accurate for downstream applications.
GravCal refines this noisy IMU prior using visual evidence from a single image, combining a prior-correction pathway that applies a learned residual rotation to $\hat{\mathbf{g}}$ with an image-only pathway that provides a fallback when the prior is unreliable.
The two pathways are blended by an adaptive gating mechanism that is jointly trained end-to-end.

\subsection{Feature Extraction}\label{sec:backbone} We use an EfficientNet-B0~\cite{tan2019efficientnet} backbone pretrained on ImageNet~\cite{deng2009imagenet} as our visual feature extractor. The final classification layer is replaced with an identity mapping, yielding a feature vector $\mathbf{f} \in \mathbb{R}^{C}$ with $C = 1280$. During training, we freeze the batch normalization layers in the backbone to preserve the pretrained statistics, and selectively unfreeze the later MBConv blocks for fine-tuning with a reduced learning rate. 

\subsection{FiLM Conditioning}\label{sec:film} To incorporate the gravity prior into the visual representation, we employ Feature-wise Linear Modulation (FiLM)~\cite{perez2018film}. A small MLP maps the prior $\hat{\mathbf{g}} \in \mathbb{R}^3$ to affine modulation parameters: \begin{equation} \boldsymbol{\gamma}, \boldsymbol{\beta} = \text{PriorMLP}(\hat{\mathbf{g}})\,, \quad \boldsymbol{\gamma}, \boldsymbol{\beta} \in \mathbb{R}^{C}\,, \label{eq:film_params} \end{equation} which modulate the backbone features as: \begin{equation} \tilde{\mathbf{f}} = \boldsymbol{\gamma} \odot \mathbf{f} + \boldsymbol{\beta}\,. \label{eq:film} \end{equation} To ensure stable training, we initialize the PriorMLP such that $\boldsymbol{\gamma} \approx \mathbf{1}$ and $\boldsymbol{\beta} \approx \mathbf{0}$ at the start, so the conditioning starts as a near-identity transformation. 

\subsection{Prior-Correction Branch}\label{sec:prior_correction} The conditioned features $\tilde{\mathbf{f}}$ are passed through an MLP head that predicts a residual rotation $(\delta_x, \delta_y) \in [-\delta_{\max}, \delta_{\max}]$, parameterized as Euler angles (pitch and yaw offsets), where $\delta_{\max} = 45^\circ$.
This range is chosen to cover the typical error magnitude of IMU-derived priors, which can reach tens of degrees under prolonged gyroscope drift or poor accelerometer conditions~\cite{mourikis2007multi,qin2018vins}. The output is constrained via the $\tanh$ activation: \begin{equation} \delta_x, \delta_y = \delta_{\max} \cdot \tanh(\mathbf{h}_{1:2})\,, \label{eq:delta} \end{equation} where $\mathbf{h} = \text{Head}(\tilde{\mathbf{f}})$. The corrected gravity estimate is obtained by applying sequential rotations $R_x(\delta_x)$ and $R_y(\delta_y)$ to the prior: \begin{equation} \mathbf{g}_{\text{corr}} = \text{normalize}\big(R_y(\delta_y)\, R_x(\delta_x)\, \hat{\mathbf{g}}\big)\,. \label{eq:g_corr} \end{equation} This branch is effective when the prior is already close to the true gravity, as it only needs to learn a small correction. 

\subsection{Image-Only Branch}\label{sec:img_branch} A separate MLP head operates on the \emph{unconditioned} features $\mathbf{f}$ (without FiLM modulation) and directly regresses a gravity direction: \begin{equation} \mathbf{g}_{\text{img}} = \text{normalize}\big(\text{ImgHead}(\mathbf{f})\big)\,. \label{eq:g_img} \end{equation} Since this branch does not depend on the prior, it can estimate gravity from any camera orientation, serving as a fallback when no reliable prior is available. 

\subsection{Adaptive Gating}\label{sec:gating} The same head that produces the residual rotation also outputs a scalar gating value $\tau \in [0, 1]$ via a sigmoid activation: \begin{equation} \tau = \sigma(\mathbf{h}_3)\,. \label{eq:tau} \end{equation} The final prediction is a convex combination of the two branches: \begin{equation} \mathbf{g}_{\text{pred}} = \text{normalize}\big(\tau \cdot \mathbf{g}_{\text{img}} + (1 - \tau) \cdot \mathbf{g}_{\text{corr}}\big)\,. \label{eq:g_pred} \end{equation} When $\tau \to 0$, the model trusts the corrected prior; when $\tau \to 1$, it relies entirely on the image-only branch. The bias of $\tau$ is initialized to $-3.0$ so that $\sigma(-3.0) \approx 0.05$, encouraging the network to initially trust the prior correction and learn to increase $\tau$ only when necessary. 

\subsection{Loss Function}\label{sec:loss}

Our training objective consists of four terms:
\begin{equation}
    \mathcal{L} = \mathcal{L}_{\text{main}} + \lambda_\delta \mathcal{L}_\delta + \lambda_\tau \mathcal{L}_\tau + \lambda_{\text{img}} \mathcal{L}_{\text{img}}\,.
    \label{eq:total_loss}
\end{equation}

\paragraph{Main loss.}
The primary supervision is the angular error between the final prediction and the ground truth:
\begin{equation}
    \mathcal{L}_{\text{main}} = \arccos\!\big(\mathbf{g}_{\text{pred}} \cdot \mathbf{g}^*\big)\,.
    \label{eq:loss_main}
\end{equation}

\paragraph{Oracle gate supervision.}
Rather than applying a fixed heuristic penalty to $\tau$, we supervise the gate with an oracle target computed from the actual quality of the prior during training.
Let $\epsilon_p = \tfrac{180}{\pi}\arccos(\hat{\mathbf{g}} \cdot \mathbf{g}^*)$ denote the angular error of the prior in degrees.
The oracle target is:
\begin{equation}
    \tau^* = \sigma\!\left(\frac{\epsilon_p - 25}{5}\right),
    \label{eq:tau_target}
\end{equation}
where $\sigma$ is the sigmoid function.
This soft target transitions smoothly from $\tau^* \approx 0$ when the prior is accurate ($\epsilon_p \le 15^\circ$, trust the correction branch) to $\tau^* \approx 1$ when the prior is severely corrupted ($\epsilon_p \ge 35°$, rely on the image branch).
The gate loss is:
\begin{equation}
    \mathcal{L}_\tau = \big(\tau - \tau^*\big)^2\,.
    \label{eq:loss_tau}
\end{equation}
Since all training sequences include real IMU data, $\tau^*$ is always computed from the true prior error.

\paragraph{Adaptive delta regularization.}
We penalize large residual rotations to keep the correction branch conservative, but relax the penalty when the prior is known to be unreliable (high $\tau^*$):
\begin{equation}
    \mathcal{L}_\delta = \frac{1}{B}\sum_{b=1}^{B} \big(1 - \tau^*_b\big)\,\big(\delta_{x,b}^2 + \delta_{y,b}^2\big)\,,
    \label{eq:loss_delta}
\end{equation}
where $\delta_x, \delta_y$ are in radians.
When the prior is accurate ($\tau^* \approx 0$), corrections must be small and precise.
When the prior is corrupted ($\tau^* \approx 1$), the image branch bears the load and large deltas from the correction branch are not penalized.

\paragraph{Adaptive image-only supervision.}
Direct supervision on $\mathbf{g}_{\text{img}}$ prevents the image-only branch from becoming dormant due to gradient attenuation through the gating mechanism.
We up-weight this supervision when the prior is poor, since the image branch must then carry the full prediction responsibility:
\begin{equation}
    \mathcal{L}_{\text{img}} = \frac{1}{B}\sum_{b=1}^{B} \big(0.5 + \tau^*_b\big)\,\arccos\!\big(\mathbf{g}_{\text{img},b} \cdot \mathbf{g}^*_b\big)\,.
    \label{eq:loss_img}
\end{equation}
The baseline weight of $0.5$ ensures constant supervision even when the prior is reliable, while the $\tau^*$ term doubles the effective supervision weight on corrupted-prior samples.

\subsection{Prior Source}\label{sec:prior_augmentation}
The gravity prior $\hat{\mathbf{g}}$ used during training comes from the IMU-based pipeline described in Section~\ref{sec:imu_prior}.
The Mahony filter output is expressed in the IMU body frame and must be rotated into the camera frame via the extrinsic rotation $R_{\text{imu}\to\text{cam}}$.
We estimate $R_{\text{imu}\to\text{cam}}$ from the data using the Orthogonal Procrustes alignment described in Section~\ref{sec:imu_prior}: the VIO-derived camera-frame gravity and the Mahony-filter IMU-frame gravity are jointly used to solve for the best-fit rotation in a single SVD step.
In practice, $R_{\text{imu}\to\text{cam}}$ is estimated per sequence using the VIO poses available from ARKit, as subtle hardware variation across devices makes a single global estimate less accurate.
No dedicated calibration target or procedure is required; any routine recording with active VIO suffices.
All training sequences have synchronized IMU streams, and so the network is trained on the real error distribution of inertial sensing rather than a synthetic approximation.

\section{Experiments}\label{sec:experiments}

\subsection{Implementation Details}\label{sec:impl}

We follow the architecture described in Sec.~\ref{sec:method}.
Images are first undistorted using the camera's intrinsic distortion model, then resized to $224 \times 224$ and normalized with ImageNet mean and standard deviation.
We use the Adam optimizer with a learning rate of $5 \times 10^{-5}$ for the prediction heads and $2 \times 10^{-6}$ for the unfrozen backbone layers.
The learning rate follows a cosine annealing schedule over 50 epochs.
Training uses a batch size of 64 with Automatic Mixed Precision (AMP).
During training, the gravity prior is the IMU-derived estimate from the Mahony complementary filter aligned to the camera frame (Section~\ref{sec:imu_prior}); all training sequences include synchronized IMU data, so real priors are used throughout.
Loss weights are set to $\lambda_\delta = 10^{-4}$, $\lambda_\tau = 0.05$, and $\lambda_{\text{img}} = 0.2$.
All experiments are conducted on a single NVIDIA RTX 4080 GPU.

\subsection{Evaluation Metric}\label{sec:metric}

For each frame, we compute the angular error in degrees between the predicted and ground-truth gravity directions:
\begin{equation}
    e_i = \arccos\!\left(\hat{\mathbf{g}}_i \cdot \mathbf{g}_i^*\right) \cdot \frac{180}{\pi}\,.
    \label{eq:angular_error}
\end{equation}
We summarize the distribution of $e_i$ across all test frames by reporting the mean, median, 90th percentile (P90), and 95th percentile (P95).

% ---------------------------------------------------------------
\subsection{Group I: Results on Our Dataset}\label{sec:results_ours}
% ---------------------------------------------------------------

\paragraph{Setup.}
We evaluate on the test split of our dataset (Sec.~\ref{sec:dataset}), comprising 19 sessions and 26\,446 images that span a broad range of camera orientations including upright, sideways, and inverted poses.
We compare our full model against GeoCalib~\cite{veicht2024geocalib}, a state-of-the-art learning-based method that estimates camera calibration parameters including the up-vector via iterative Levenberg--Marquardt optimization.
GeoCalib operates purely on visual features with no IMU input, providing a strong image-only reference point.
We also include two standard geometry baselines: \textit{Assume Upright}, which always predicts $\hat{\mathbf{g}}=(0,1,0)^{\top}$ in the camera frame, and a \textit{VP Estimator} that recovers the vertical vanishing point from LSD line segments~\cite{vongioi2010lsd} via RANSAC.

\begin{figure}[t]
    \centering
    \scalebox{0.85}{
    \includegraphics[width=\linewidth]{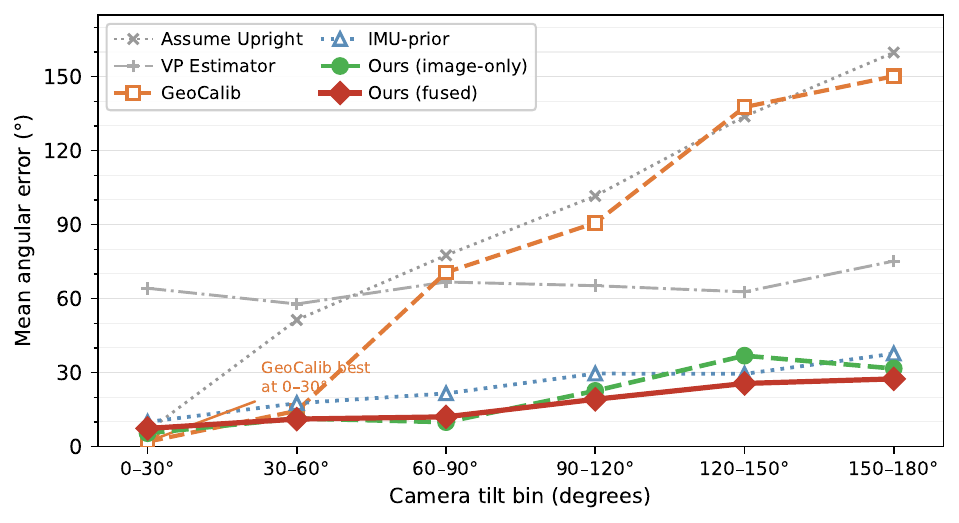}
    }
    \caption{Mean angular error (degrees $\downarrow$) on our test set, broken down by camera tilt angle in $30^\circ$ increments. GeoCalib excels near upright but collapses beyond $30^\circ$; the fused model benefits from the inertial prior and leads in four of six bins. Full numerical values are provided in the supplementary material.}
    \label{fig:tilt_breakdown}
\end{figure}

\paragraph{Results.}
Table~\ref{tab:results_ours} reports angular error statistics on our test set.
GeoCalib, while competitive on near-upright frames, degrades sharply for tilted and inverted camera poses due to its implicit assumption of a canonical upright scene; as a result, its overall mean on our orientation-diverse test set coincidentally matches Assume Upright, a pattern made explicit in Figure~\ref{fig:tilt_breakdown}.
The IMU-prior alone already outperforms all geometric and image-only baselines on the full test set, confirming that inertial sensing provides a strong directional cue even without visual refinement.
Our image-only branch surpasses GeoCalib overall, demonstrating that training on a diverse orientation distribution generalises better than methods optimised for upright imagery.
The fused model achieves the best mean error ($14.24^\circ$) and the lowest P95 ($49.18^\circ$), reflecting stronger tail robustness; the image-only branch leads on median and P90, where the prior offers diminishing returns over well-trained visual features.

\begin{table}[t]
    \centering
    \caption{Results on our test set. Angular error in degrees ($\downarrow$ lower is better).}
    \label{tab:results_ours}
    \setlength{\tabcolsep}{5pt}
    \small

    \scalebox{0.95}{
    \begin{tabular}{lcccc}
        \toprule
        Method & Mean & Median & P90 & P95 \\
        \midrule
        Assume Upright                     & $76.05^\circ$ & $80.53^\circ$ & $126.88^\circ$ & $144.68^\circ$ \\
        VP Estimator~\cite{vongioi2010lsd}  & $64.77^\circ$ & $67.72^\circ$ & $85.51^\circ$ & $87.84^\circ$ \\
        GeoCalib~\cite{veicht2024geocalib} & $76.05^\circ$ & $80.52^\circ$ & $126.88^\circ$ & $144.68^\circ$ \\
        IMU-prior~\cite{mahony2008nonlinear} & $22.02^\circ$ & $13.89^\circ$ & $52.47^\circ$ & $69.33^\circ$ \\
        \midrule
        Ours (IMU-only)                    & $22.94^\circ$ & $15.61^\circ$ & $51.09^\circ$ & $67.36^\circ$ \\
        Ours (image-only)                  & $14.81^\circ$ & \cellcolor{gray!20}$7.15^\circ$  & \cellcolor{gray!20}$29.31^\circ$ & $63.05^\circ$ \\
        \textbf{Ours (fused)}              & \cellcolor{gray!20}$14.24^\circ$ & $7.60^\circ$ & $32.68^\circ$ & \cellcolor{gray!20}$49.18^\circ$ \\
        \bottomrule
    \end{tabular}
    }
\end{table}

\paragraph{Orientation breakdown.}
To understand how each method behaves across different camera orientations, we partition the test set into six $30^\circ$ tilt bins.
Figure~\ref{fig:tilt_breakdown} plots mean angular error per bin.
GeoCalib achieves the lowest error in the near-upright bin ($0^\circ$--$30^\circ$, $1.98^\circ$), confirming its strength on images close to the canonical upright assumption it was trained on; however, performance degrades sharply beyond $30^\circ$, reaching $70.62^\circ$ at $60^\circ$--$90^\circ$ and exceeding $150^\circ$ at $150^\circ$--$180^\circ$.
Assume Upright degrades monotonically with tilt, as expected.
The IMU-prior and IMU-only variants remain stable across all tilt bins, as inertial sensing is orientation-agnostic.
Our image-only branch leads at $60^\circ$--$90^\circ$ ($9.9^\circ$), where structured visual cues are still abundant and the inertial prior adds limited benefit.
The fused model achieves the best error in four of six bins---all regimes beyond $30^\circ$ except $60^\circ$--$90^\circ$---with the largest absolute gain over image-only in the heavily tilted bins ($25.59^\circ$ vs.\ $36.86^\circ$ at $120^\circ$--$150^\circ$; $27.45^\circ$ vs.\ $31.65^\circ$ at $150^\circ$--$180^\circ$), where horizon-based visual cues become unreliable and the inertial prior is most valuable.

% ---------------------------------------------------------------
\subsection{Group II: Cross-Dataset Evaluation and Ablations}\label{sec:results_cross}
% ---------------------------------------------------------------

\paragraph{Setup.}
To assess generalization, we evaluate on three external benchmarks:
\textbf{EuRoC}~\cite{burri2016euroc}, a MAV benchmark with motion-capture ground truth recorded in indoor machine-hall and Vicon-room environments; the aggressive flight dynamics induce large linear accelerations that frequently corrupt the accelerometer-based gravity estimate, stressing the IMU prior;
\textbf{UZH-FPV}~\cite{Delmerico19icra}, a first-person-view drone racing dataset with high-agility trajectories and ground truth from a Vicon system, providing an extreme test case where the vehicle is rarely in a quasi-static state;
and \textbf{TUM-VI}~\cite{schubert2018tumvi}, a visual-inertial benchmark with motion-capture ground truth covering both handheld indoor and outdoor sequences.
We evaluate on three sequences spanning different environments: \texttt{corridor1}, \texttt{corridor2}, and \texttt{magistrale6}; results are averaged across the three sequences.

To disentangle the contributions of the two input modalities, we compare three variants of our model:
\textit{IMU-only} uses only the Mahony-filter gravity prior with no image correction ($\tau = 0$ fixed);
\textit{image-only} ignores the IMU and predicts gravity purely from visual features ($\tau = 1$ fixed);
\textit{fused} is the full model with adaptive gating.
All three variants are first evaluated \emph{zero-shot}, i.e., trained exclusively on our dataset with no samples from the target benchmark.

For UZH-FPV, we additionally evaluate two fine-tuned variants: \textit{image-only (FT)} and \textit{fused (FT)}, where the respective pretrained models are fine-tuned for 2 epochs on 500 randomly sampled UZH-FPV frames held out from the test sequences.
This lightweight adaptation protocol tests whether the model can quickly absorb domain-specific visual cues (\eg propeller occlusions, extreme motion blur) that are absent from the training distribution, and whether IMU fusion still provides additional benefit after domain adaptation.

\paragraph{Results.}
Table~\ref{tab:results_cross} reports mean angular error across all three benchmarks.
The IMU-only baseline performs well on EuRoC sequences with relatively low dynamics but degrades significantly on UZH-FPV, where sustained thrust prevents the accelerometer from providing a reliable gravity reference.
The image-only variant is stable across platforms but lacks the precision achievable when a reliable inertial prior is available.
The zero-shot fused model consistently outperforms both ablations, confirming that adaptive gating correctly up-weights the more reliable modality in each scenario.

On UZH-FPV, all zero-shot variants struggle due to the large domain gap between FPV racing imagery and our training data.
Fine-tuning with only 500 domain images dramatically reduces error for both variants: image-only drops from 31.59$^\circ$ to 12.69$^\circ$, and fused drops from $29.56^\circ$ to 10.58$^\circ$.
The persistent gap between fused (FT) and image-only (FT) demonstrates that IMU fusion continues to provide complementary information even after domain adaptation.

\begin{table}[t]
    \centering
    \caption{Cross-dataset mean angular error (degrees $\downarrow$). Zero-shot variants are trained on our dataset only; \textit{(FT)} variants are additionally fine-tuned on 500 UZH-FPV frames. Best per column \colorbox{gray!20}{highlighted}.}
    \label{tab:results_cross}
    \setlength{\tabcolsep}{5pt}
    \small
    \begin{tabular}{lcccc}
        \toprule
        Method & EuRoC & UZH-FPV & TUM VI\\
        \midrule
        IMU Prior~\cite{mahony2008nonlinear} & $5.08^\circ$ & \cellcolor{gray!20}$29.4^\circ$  & $13.48^\circ$  \\
        Ours (IMU-only)            & $5.10^\circ$ & $29.97^\circ$  & $13.56^\circ$  \\
        Ours (image-only)          & $5.14^\circ$ & $31.59^\circ$  & $27.86^\circ$  \\
        Ours (fused)               & \cellcolor{gray!20}$4.93^\circ$ & $29.56^\circ$  & \cellcolor{gray!20}$12.52^\circ$ \\
        \midrule
        Ours (image-only, FT)              & ---  & $12.69^\circ$ & ---    \\
        \textbf{Ours (fused, FT)}          & ---  & \cellcolor{gray!20}$10.58^\circ$ & --- \\
        \bottomrule
    \end{tabular}
\end{table}

% ---------------------------------------------------------------
\subsection{Group III: Gating Confidence Diagnosis}\label{sec:gate_diagnosis}
% ---------------------------------------------------------------

\paragraph{Setup.}
A key component of our model is the adaptive gate $\tau \in [0,1]$, which controls the relative contribution of the IMU prior and the image branch (Sec.~\ref{sec:method}).
To verify that the gate behaves meaningfully rather than collapsing to a degenerate constant, we analyze its predicted values against two independent quality proxies that are \emph{not} used during training:
(i)~the \textit{IMU prior error} $\epsilon_{\text{imu}}$, the angular difference between the Mahony-filter estimate and ground-truth gravity; and
(ii)~the \textit{accelerometer non-gravity ratio} $r = \|\mathbf{a}\| / g - 1$, which measures the fraction of linear acceleration contaminating the inertial measurement.

\paragraph{Results.}
Table~\ref{tab:tau_vs_prior_error} reports the mean predicted gate $\tau$ binned by the IMU prior error $\epsilon_{\text{imu}}$.
Since the oracle target is $\tau^* = \sigma((\epsilon_p - 25^\circ)/5^\circ)$, the gate is expected to rise steeply around the $20^\circ$--$30^\circ$ range.
The data confirm this: $\tau$ increases from $0.52$ at low prior errors ($<10^\circ$) to a peak of $0.65$ in the $20^\circ$--$30^\circ$ bin, then gradually decreases for larger errors.
The slight decrease beyond $30^\circ$ is consistent with the training data distribution: the prior error distribution peaks around $38^\circ$ and samples with errors beyond $60^\circ$ are rare, since complete IMU failure is uncommon in practice.
The model therefore has limited supervision signal for extreme prior errors and reverts towards a moderate $\tau$.

The more informative diagnostic is the correlation with the \emph{accelerometer non-gravity ratio} $r$, which is never seen during training.
Table~\ref{tab:gate_diag} shows that $\tau$ rises from $0.57$ at $r < 0.05$ to $0.72$ at $r > 0.50$, confirming that the gate generalizes its learned behavior to a physically meaningful but unsupervised signal.

\begin{table}[t]
    \centering
    \caption{Mean predicted gate $\tau$ binned by IMU prior error $\epsilon_{\text{imu}}$. The gate peaks near the oracle sigmoid center ($25^\circ$); the plateau beyond $30^\circ$ reflects the scarcity of high-error training samples.}
    \label{tab:tau_vs_prior_error}
    \setlength{\tabcolsep}{5pt}
    \small
    \begin{tabular}{lcccccc}
        \toprule
        Prior error $\epsilon_{\text{imu}}$ & $<10^\circ$ & $10^\circ$--$20^\circ$ & $20^\circ$--$30^\circ$ & $30^\circ$--$45^\circ$ & $45^\circ$--$60^\circ$ & $>60^\circ$ \\
        \midrule
        Mean $\tau$ & 0.52 & 0.63 & \cellcolor{gray!20}0.65 & 0.62 & 0.61 & 0.59 \\
        \bottomrule
    \end{tabular}
\end{table}

\begin{table}[t]
    \centering
    \caption{Mean predicted gate $\tau$ per accelerometer non-gravity ratio bin. Higher $r$ indicates larger dynamic acceleration corrupting the IMU prior; a reliable gate should increase $\tau$ (lean towards image branch) as $r$ grows.}
    \label{tab:gate_diag}
    \setlength{\tabcolsep}{6pt}
    \small
    \begin{tabular}{lccccc}
        \toprule
        Non-gravity ratio $r$ & $<0.05$ & $0.05$--$0.10$ & $0.10$--$0.20$ & $0.20$--$0.50$ & $>0.50$ \\
        \midrule
        Mean $\tau$           & 0.57    & 0.59           & 0.57           & 0.61           & \cellcolor{gray!20}0.72    \\
        \bottomrule
    \end{tabular}
\end{table}

% ---------------------------------------------------------------
\subsection{Group IV: Runtime and Latency}\label{sec:latency}
% ---------------------------------------------------------------

\paragraph{Setup.}
We measure per-frame inference latency on an iPhone 13 Pro Max (batch size 1, mean over 1000 frames) for two stages: (1)~\textit{undistortion and resizing}, and (2)~\textit{EfficientNet-B0 backbone inference}.

\paragraph{Results.}
We fuse undistortion and resizing into a single remapping step that directly produces a $224 \times 224$ rectified crop from a $1920 \times 1440$ input, taking ${\approx}1$\,ms per frame.
Since the remap table depends only on the camera intrinsics and is fixed at deployment time, it can be precomputed once and stored, adding no per-frame overhead beyond the lookup itself. The EfficientNet-B0 backbone runs in ${\approx}3$\,ms.
Overall, the full GravCal pipeline runs in 4\,ms per frame at batch size 1, well within real-time budgets for mobile deployment.

\subsection{Future Work}
GravCal relies on an EfficientNet-B0 backbone,
%, which balances efficiency and capacity but may still limit representational power for subtle, fine-grained orientation cues.
future work could explore larger backbone architectures such as Vision Transformers (ViT)~\cite{dosovitskiy2020image}, which may better capture global geometric structure relevant for gravity estimation.
Additionally, the IMU-to-camera extrinsic rotation $R_{\text{imu}\to\text{cam}}$ is currently estimated per sequence via Orthogonal Procrustes using available VIO poses; an interesting direction is to learn this alignment jointly with the gravity prediction, removing the need for any separate calibration step.
Finally, while GravCal handles arbitrary orientations, performance on near-inverted poses ($120^\circ$--$180^\circ$) remains lower than on other bins due to limited training data in this regime, which we aim to address in future data collection efforts.
%We presented GravCal, a feedforward calibrator for single-image gravity prior calibration. Given an IMU-derived gravity prior---which is directionally informative but corrupted by gyroscope drift and accelerometer noise---and one RGB image, GravCal predicts a residual correction on $S^2$ together with a confidence score, without any assumption on camera orientation. We also introduced a large-scale dataset of over 148K images with dual annotation: VIO-derived ground-truth gravity and Mahony-filter IMU priors, covering a broad range of orientations including heavily tilted and near-inverted poses.

%Experiments demonstrate that GravCal reduces the mean tilt error from $22.02^\circ$ (IMU prior alone) to $14.24^\circ$, with particularly large gains on tail cases where the prior is most corrupted. The predicted gate $\tau$ correlates with IMU prior quality---rising with accelerometer non-gravity ratio and peaking near the oracle sigmoid center---confirming that it provides a meaningful per-sample reliability signal beyond its training supervision.

\section{Conclusion}\label{sec:conclusion}

GravCal shows that noisy inertial gravity priors need not be blindly trusted or discarded: they can be calibrated from a single image, reliably and efficiently, at inference time. By delivering strong gains across diverse scenes and camera orientations, our results suggest a broader direction beyond this task itself: perception systems should not merely fuse learned predictions with physical signals, but actively assess and repair unreliable sensor priors before they propagate downstream. We believe this work helps move visual-inertial perception toward more robust, confidence-aware, and deployment-ready AR and robotics systems.

\bibliography{main}
%\appendix
%\newpage

%\addcontentsline{toc}{section}{Appendix}

%\begin{center}
%{\Large\bfseries\color{AweAIblue} Contents of Appendix}
%\end{center}

%\startcontents[appendix]
%\printcontents[appendix]{}{1}{\setcounter{tocdepth}{2}}

\end{document}